\documentclass[twocolumn]{svjour3}

\usepackage{times}
\usepackage{epsfig}
\usepackage{graphicx}
\usepackage{amsmath}
\usepackage{amssymb}
\usepackage{caption}
\usepackage{multirow}
\usepackage{floatrow}
\usepackage{hyperref}
\usepackage{color}

\definecolor{ForestGreen}{rgb}{0.0, 0.27, 0.13}
\definecolor{BrickRed}{rgb}{0.8, 0.25, 0.33}

\newcommand{\no}{-}
\newcommand{\yes}{yes}
\newcommand{\rebuttal}[1]{{ \color{black} #1 }}
\newcommand{\revision}[1]{{ \color{black} #1 }}

\begin{document}

\title{Every Moment Counts: Dense Detailed Labeling of Actions in Complex Videos}

\author{Serena Yeung \and Olga Russakovsky 
\and Ning Jin \and Mykhaylo Andriluka \and
Greg Mori \and Li Fei-Fei}


\institute{
S. Yeung \at
Stanford University, Stanford, CA, USA \\
\email{serena@cs.stanford.edu}
\and
O. Russakovsky \at
Carnegie Mellon University, Pittsburgh, PA, USA \\
Stanford University, Stanford, CA, USA
\and
N. Jin \at
Stanford University, Stanford, CA, USA
\and
M. Andriluka \at
Stanford University, Stanford, CA, USA \\
Max Planck Institute for Informatics, Saarbr\"ucken, Germany
\and
G. Mori \at
Simon Fraser University, Burnaby, BC, Canada
\and
L. Fei-Fei \at
Stanford University, Stanford, CA, USA
}

\maketitle

\begin{abstract}
   Every moment counts in action recognition.  A comprehensive understanding of human activity in video requires labeling every frame according to the actions occurring, placing multiple labels densely over a video sequence.  To study this problem we extend the existing THUMOS dataset and introduce MultiTHUMOS, a new dataset of dense labels over unconstrained internet videos.  Modeling multiple, dense labels benefits from temporal relations within and across classes.  We define a novel variant of long short-term memory (LSTM) deep networks for modeling these temporal relations via multiple input and output connections.  We show that this model improves action labeling accuracy and further enables deeper understanding tasks ranging from structured retrieval to action prediction.
\end{abstract}


\section{Introduction}

\begin{figure}
    \centering
    \includegraphics[width=1\linewidth]{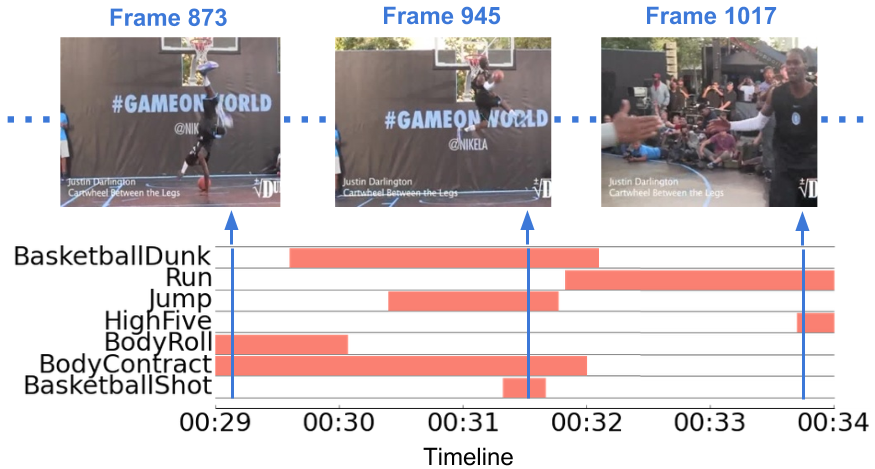}
    \caption{In most internet videos there are multiple simultaneous human actions. Here, we show a concrete example from a basketball video to illustrate our target problem of dense detailed multi-label action understanding.}
    \label{fig:pullfig}
\end{figure}

Humans are great at multi-tasking: they can be walking while talking on the phone while holding a cup of coffee. Further, human action is continual, and every minute is filled with potential labeled actions (Figure~\ref{fig:pullfig}). However, most work on human action recognition in video focuses on recognizing discrete instances or single actions at a time: for example, which sport~\cite{karpathy2014large} or which single cooking activity~\cite{rohrbach2012database} is taking place. We argue this setup is fundamentally limiting. First, a single description is often insufficient to fully describe a person's activity. Second, operating in a single-action regime largely ignores the intuition that actions are intricately connected. A person that is running and then jumping is likely to be simultaneously doing a sport such as basketball or long jump; a nurse that is taking a patient's blood pressure and looking worried is likely to call a doctor as her next action. In this work, we go beyond the standard one-label paradigm to dense, detailed, multilabel understanding of human actions in videos. 

There are two key steps on the path to tackling detailed multilabel human action understanding: (1) finding the right dataset and (2) developing an appropriate model.  In this paper we present work in both dimensions.

The desiderata for a video dataset include the following: video clips need to be long enough to capture multiple consecutive actions, multiple simultaneous actions need to be annotated, and labeling must be  dense with thorough coverage of action extents.
Video annotation is very time-consuming and expensive, and to the best of our knowledge no such dataset currently exists. UCF101~\cite{soomro2012ucf101},  HMDB51~\cite{kuehne2011hmdb}, and Sports1M~\cite{karpathy2014large} are common challenging action recognition datasets. However, each video is associated with non-localized labels (Sports1M), and the videos in UCF101 and HMDB51 are further temporally clipped around the action.  MPII Cooking~\cite{rohrbach2012database} and Breakfast~\cite{kuehne2014language} datasets contain long untrimmed video sequences with multiple sequential actions but still only one label per frame; further, they are restricted to closed-world kitchen environments. THUMOS ~\cite{THUMOS14} contains long untrimmed videos but most videos ($85\%$) only contain a single action class. 

To overcome these problems, we introduce a new action detection dataset called MultiTHUMOS, significantly extending the annotations on 413 videos (30 hours) of THUMOS action detection dataset. First, MultiTHUMOS allows for an in-depth study of simultaneous human action in video: it extends THUMOS from $20$ action classes with $0.3$ labels per frame to $65$ classes and $1.5$ labels per frame. Second, MultiTHUMOS allows for a thorough study of the temporal interaction between consecutive actions: the average number of distinct action categories in a video is $10.5$ (compared to $1.1$ in THUMOS). Going further, MultiTHUMOS lends itself to studying intricate relationships between action labels: the $45$ new annotated classes include relationships such as hierarchical (e.g., more general Throw or PoleVault and more specific BasketballShot or PoleVaultPlantPole) and fine-grained (e.g., Guard versus Block or Dribble versus Pass in basketball). Figure~\ref{fig:pullfig} shows an example of our dense multilabel annotation.

Reasoning about multiple, dense labels on video requires models capable of incorporating temporal dependencies.  A large set of techniques exist for modeling temporal structure, such as hidden Markov models (HMMs), dynamic time warping, and their variants.  Recent action recognition literature has used recurrent neural networks known as Long Short Term Memory (LSTM) for action recognition in videos \cite{donahue2014long}. We introduce MultiLSTM, a new LSTM-based model targeting dense, multilabel action analysis. Taking advantage of the fact that more than $45\%$ of frames in MultiTHUMOS have 2 or more labels, the model can learn dependencies between actions in nearby frames and between actions in the same frame, which allows it to subsequently perform dense multilabel temporal action detection on unseen videos. 

In summary, our contributions are:
\begin{enumerate}
\item We introduce MultiTHUMOS, a new large-scale dataset of dense, multilabel action annotations in temporally \\ untrimmed videos, and
\item We introduce MultiLSTM, a new recurrent model based on an LSTM that features temporally-extended input and output connections. 
\end{enumerate}
Our experiments demonstrate improved performance of MultiLSTM relative to a plain LSTM baseline on our dense, multilabel action detection benchmark.



\section {Related Work}

Visual analysis of human activity has a long history in computer vision research.  Thorough surveys of the literature include Poppe~\cite{Poppe10} and Weinland et al.~\cite{WeinlandRB10}.  Here we review recent work relevant to dense labeling of videos.

\subsection{Datasets} Research focus is closely intertwined with dataset creation and availability.  The KTH~\cite{SchuldtLC04} and Weizmann~\cite{Blank05} datasets were catalysts for a body of work.  This era focused on recognizing individual human actions, based on datasets consisting of an individual human imaged against a generally stationary background.  In subsequent years, the attention of the community moved towards more challenging tasks.  Benchmarks based on surveillance video were developed for crowded scenes, such as the TRECVID Surveillance Event Detection~\cite{OverAMFKSQ11}.  Interactions between humans or humans and objects~\cite{ryoo09,Oh2011} have been studied.

Another line of work has shifted toward analyzing ``unconstrained" internet video.  Datasets in this line present challenges in the level of background clutter present in the videos.  The Hollywood (HOHA)~\cite{marszalek09}, HMDB~\cite{kuehne2011hmdb}, UCF 101~\cite{soomro2012ucf101}, ActivityNet~\cite{caba2015activitynet}, and THUMOS ~\cite{THUMOS14} datasets exemplify this trend.  Task direction has also moved toward a retrieval setting, finding a (small) set of videos from a large background collection, including datasets such as TRECVID MED~\cite{OverAMFKSQ11} and Sports 1M~\cite{karpathy2014large}.

While the push toward unconstrained internet video is positive in terms of the difficulty of this task, it has moved focus away from human action toward identifying scene context.  Discriminating diving versus gymnastics largely involves determining the scene of the event.  The MPII Cooking dataset \cite{rohrbach2012database} and Breakfast dataset \cite{kuehne2014language} refocus efforts toward human action within restricted action domains (Table~\ref{table:compare_datasets}).  The MultiTHUMOS dataset we propose shares commonalities with this line, but emphasizes generality of video, multiple labels per frame, and a broad set of general to specific actions.

\subsection{Deep learning for video} In common with object recognition, hand-crafted features for video analysis are giving way to deep convolutional feature learning strategies.  The best hand-crafted features, the dense trajectories of Wang et al.~\cite{WangKSL11}, achieve excellent results on benchmark action recognition datasets.  However, recent work has shown superior results by learning video features (often combined with dense trajectories).  Simonyan and Zisserman~\cite{simonyan2014two} present a two-stream convolutional architecture utilizing both image and optical flow data as input sources.  Zha et al.~\cite{zha2015exploiting} examine aggregation strategies for combining deep learned image-based features for each frame, obtaining impressive results on TRECVID MED retrieval.  Karpathy et al.~\cite{karpathy2014large} and Tran et al.~\cite{TranBFTP15} learn spatio-temporal filters in a deep network and apply them to a variety of human action understanding tasks.  Mansimov et al.~\cite{mansimov2015initialization} consider methods for incorporating ImageNet training data to assist in initializing model parameters for learning spatio-temporal features.  Wang et al.~\cite{wang2015temporal} study temporal pooling strategies, specifically focused on classification in variable-length input videos.

\begin{table}
\scriptsize
\begin{center}
\begin{tabular}{|l|c|c|c|c|}
\hline
& Detection & Untrimmed & Open-world & Multilabel\\
\hline
UCF101~\cite{soomro2012ucf101} & \no & \no & \yes & \no \\
HMDB51~\cite{kuehne2011hmdb} & \no & \no & \yes & \no \\
Sports1M~\cite{karpathy2014large} & \no & \yes & \yes & \no\\
Cooking~\cite{rohrbach2012database} & \yes & \yes & \no & \no \\
Breakfast~\cite{kuehne2014language} & \yes & \yes & \no & \no \\
THUMOS~\cite{THUMOS14} & \yes & \yes & \yes & \no \\
\hline
MultiTHUMOS & \yes & \yes & \yes & \yes \\
\hline
\end{tabular}
\end{center}
\caption{Our MultiTHUMOS dataset overcomes many limitations of previous datasets.}
\label{table:compare_datasets}
\end{table}

\subsection{Temporal models for video}  Constructing models of the temporal evolution of actions has deep roots in the literature.  Early work includes Yamato et al.~\cite{yamato92_cvpr}, using hidden Markov models (HMMs) for latent action state spaces.  Lv and Nevatia~\cite{LvN07} represented actions as a sequence of synthetic 2D human poses rendered from different view points. Constraints on transitions between key poses are represented using a state diagram called an ``Action Net" which is constructed based on the order of key poses of an action.  Shi et al.~\cite{ShiCWS11} proposes a semi-Markov model to segment a sequence temporally and label segments with an action class.  Tang et al.~\cite{TangCVPR12} extend HMMs to model the duration of each hidden state in addition to the transition parameters of hidden states.

Temporal feature aggregation is another common strategy for handling video data.  Pooling models include aggregating over space and time, early and late fusion strategies, and temporal localization~\cite{tong2014lamp,myers2014evaluating,oh2014multimedia}.

Discriminative models include those based on latent SVMs over key poses and action grammars~\cite{niebles10_eccv,VahdatGRM11,PirsiavashR14}.  A recent set of 
papers has deployed deep models using long short-term memory (LSTM) models~\cite{HochreiterS97} for video analysis~\cite{donahue2014long,ng2015beyond,srivastava2015unsupervised,yao2015video}.  These papers have shown promising results applying LSTMs for tasks including video classification and sentence generation.  In contrast, we develop a novel LSTM that performs spatial input aggregation and output modeling for dense labeling output.

\subsection{Action detection} Beyond assigning a single label to a whole video, the task of action detection localizes this action within the video sequence.  An example of canonical work in this vein is Ke et al.~\cite{ke07_iccv}.  More recent work extended latent SVMs to spatio-temporal action detection and localization~\cite{tian2013spatiotemporal,lan2011discriminative}.  Rohrbach et al.~\cite{rohrbach2015recognizing} detect cooking actions using hand-centric features accounting for human pose variation.  Ni et al.~\cite{ni2014multiple} similarly utilize hand-centric features on the MPII Cooking dataset, but focus on multiple levels of action granularity.  Gkioxari and Malik~\cite{gkioxari2014finding} train SVMs for actions on top of deep learned features, and further link them in time for spatio-temporal action detection. In contrast, we address the task of dense multilabel action detection.

\subsection{Attention-based models} Seminal work on computational spatial attention models for images was done by Itti et al.~\cite{ittiKN98}.  Recent action analysis work utilizing attention includes Shapovalova et al.~\cite{shapovalova2013action} who use eye-gaze data to drive action detection and localization.
Xu et al.~\cite{xu2015show} use visual attention to assist in caption generation.  Yao et al.~\cite{yao2015video} develop an LSTM for video caption generation with soft temporal attention.  Our method builds on these directions, using an attention-based input temporal context for dense action labeling.

\section {The MultiTHUMOS Dataset}
\label{sec:dataset}

Research on detailed, multilabel action understanding requires a dataset of untrimmed, densely labeled videos.  However, we are not aware of any existing dataset that fits these requirements. THUMOS~\cite{THUMOS14} is untrimmed but contains on average only a single distinct action labeled per video.  MPII Cooking~\cite{rohrbach2012database} and Breakfast~\cite{kuehne2014language} datasets have labels of sequential actions, but contain only a single label per frame and are further captured in closed-world settings of a single or small set of kitchens (Table~\ref{table:compare_datasets}).

\begin{figure}
\centering
\includegraphics[width=\linewidth]{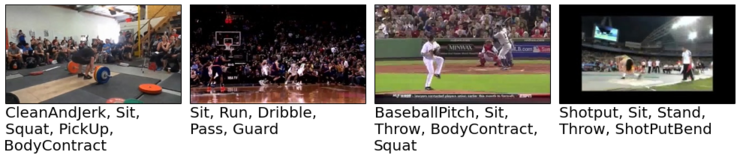} \\
\includegraphics[width=\linewidth]{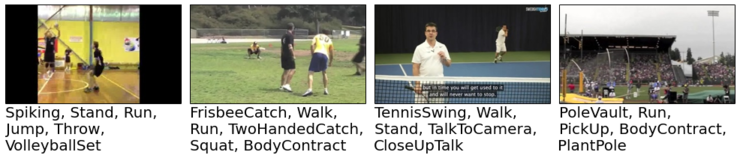} \\
\includegraphics[width=\linewidth]{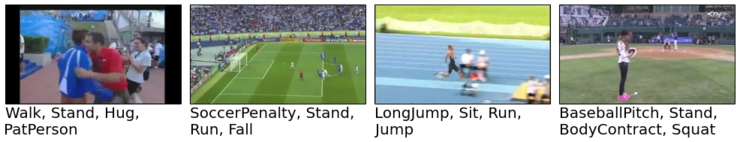} \\
 \includegraphics[width=\linewidth]{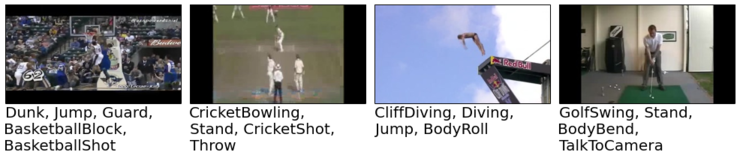} \\
    \caption{Our MultiTHUMOS dataset contains multiple action annotations per frame.}
    \label{fig:denselabels}
\end{figure}

\begin{figure}
\centering
\begin{tabular}{@{}c@{\hskip 0.2in}c@{}}
\includegraphics[width=0.47\linewidth]{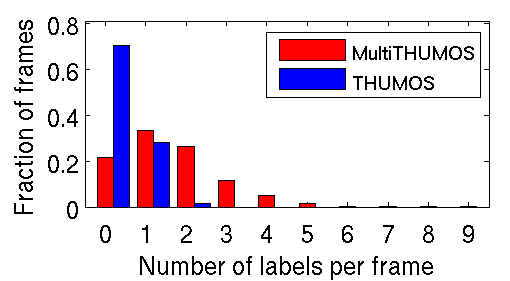} &
\includegraphics[width=0.47\linewidth]{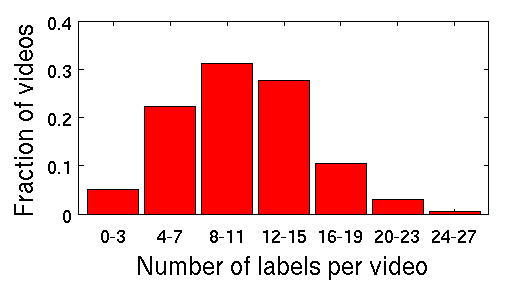}
\end{tabular}
\caption{\emph{Left.} MultiTHUMOS has significantly more labels per frame than THUMOS~\cite{THUMOS14} ($1.5$ in MultiTHUMOS versus $0.3$ in THUMOS). \emph{Right.}  Additionally, MultiTHUMOS contains up to 25 action labels per video compared to $\leq 3$ labels in THUMOS. }
\label{fig:hist-dense}
\end{figure}

\begin{figure*}
\centering
\includegraphics[width=\linewidth]{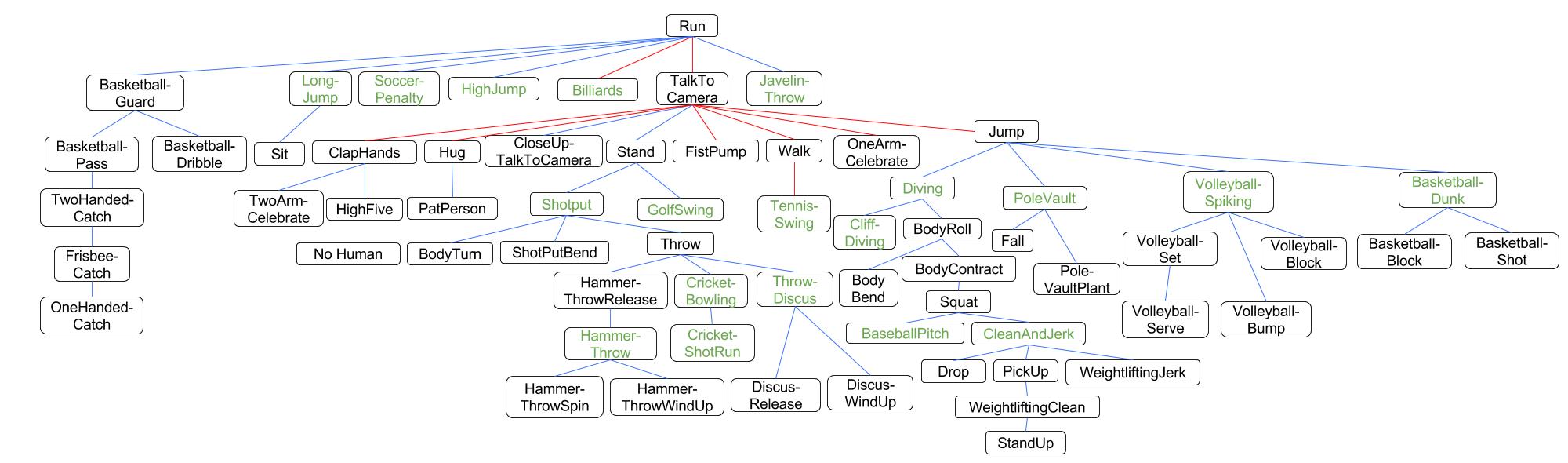} \\
\caption{We use the method of~\cite{Choi_cvpr10} to learn the relationships between the 65 MultiTHUMOS classes based on per-frame annotations. Blue (red) means positive (negative) correlation. The 20 original THUMOS classes are in green.}
\label{fig:hierarchy}
\end{figure*}

\begin{figure}
\centering
\begin{tabular}{@{}c@{\hskip 0.2in}c@{}}
\includegraphics[width=0.47\linewidth]{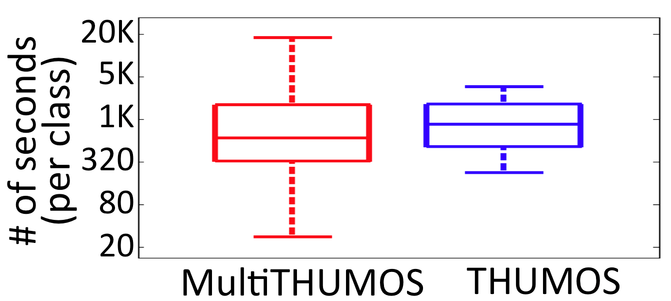} &
\includegraphics[width=0.47\linewidth]{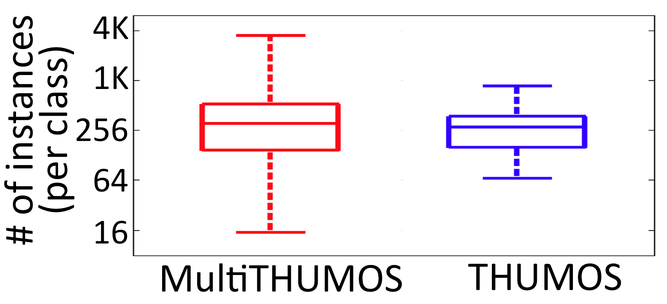} \\ 
\end{tabular}
\caption{MultiTHUMOS has a wider range of number of per-class frames and instances (contiguous sequences of a label) annotated than THUMOS. Some  action classes like Stand or Run have up to 3.5K instances (up to 18K seconds, or 5.0 hours); others like VolleyballSet or Hug have only 15 and 46 instances (27 and 50 secs) respectively. 
}
\label{fig:hist-training-data}
\end{figure}

\begin{figure*}
    \centering
    Action \#30/65: Hug
  \begin{tabular}{c}
    \includegraphics[width=\linewidth]{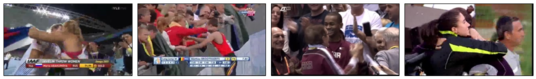} \\
    \includegraphics[width=\linewidth]{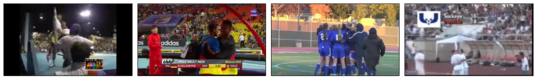} \\
\includegraphics[width=\linewidth]{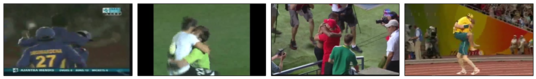} \\
    \includegraphics[width=\linewidth]{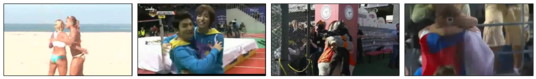}
    \end{tabular}
  Action \#46/65: BasketballDribble
     \begin{tabular}{c}
    \includegraphics[width=\linewidth]{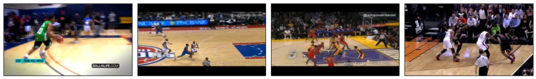} \\
    \includegraphics[width=\linewidth]{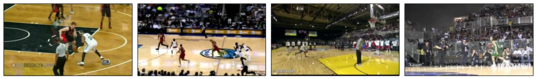} \\
    \includegraphics[width=\linewidth]{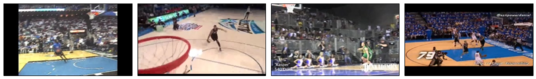} \\
    \includegraphics[width=\linewidth]{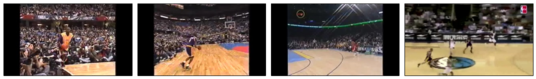} \\
    \end{tabular}
    \caption{Our MultiTHUMOS dataset is very challenging due to high intra-class variation.}
    \label{fig:intra-class-variation}
\end{figure*}


To address the limitations of previous datasets, we introduce a new dataset called MultiTHUMOS\footnote{The dataset is available for download at \url{http://ai.stanford.edu/~syyeung/everymoment.html}.}. MultiTHUMOS contains dense, multilabel, frame-level action annotations (Figure~\ref{fig:denselabels}) for 30 hours across 400 videos in the THUMOS '14 action detection dataset \revision{(referred to hereafter as THUMOS)}.  In particular, all videos in the ``Validation Data" and ``Test Data" sets were labeled.  THUMOS training data consists of 3 sets of videos: temporally clipped ``Training Data", temporally untrimmed ``Validation Data" with temporal annotations, and temporally untrimmed ``Background Data" with no temporal annotations.  Test data consists of temporally untrimmed ``Test Data" with temporal annotations.  We annotated all video sets originally including temporal annotations, i.e. ``Validation Data" and ``Test Data".

\revision{Annotations were collected in collaboration with Datatang\footnote{http://factory.datatang.com/en/}, a commercial data annotation service. Workers were provided with the name of an action, a brief (up to 1 sentence) description, and 2 annotation examples, and asked to annotate the start and end frame of the action in the videos.  An action was annotated if it occurred anywhere in the frame. A single worker was used to annotate each video since the workers are employees of the company, and a second worker verified each annotation as part of Datatang's quality control process after annotation.}


 In total, we collected $32,325$ annotations of 45 action classes, bringing the total number of annotations from $6,365$ over 20 classes in THUMOS to $38,690$ over 65 classes in MultiTHUMOS. \revision{The classes were selected to have a diversity of length, to include hierarchical, hierarchical within a sport, and fine-grained categories, and to include both sport specific and non-sport specific categories.} The action classes are described in more detail below. Importantly, it is not just the scale of the dataset that has increased. The \emph{density} of annotations increased from $0.3$ to $1.5$ labels per frame on average and from $1.1$ to $10.5$ action classes per video. The availability of such densely labeled videos allows research on interaction between actions that was previously impossible with more sparsely labeled datasets. The maximum number of actions per frame increased from $2$ in THUMOS to $9$ MultiTHUMOS, and the maximum number of actions per video increased from $3$ in THUMOS to $25$ in MultiTHUMOS. Figure~\ref{fig:hist-dense} shows the full distribution of annotation density.
 
Using these dense multilabel video annotations, we are able to learn and visualize the relationships between actions. The co-occurrence hierarchy of object classes in images based on mutual information of object annotations was learned by Choi et al.~\cite{Choi_cvpr10}; we adapt their method to per-frame action annotations in video. Figure~\ref{fig:hierarchy} shows the resulting action hierarchy. Classes such as squat and body contract frequently co-occur; in contrast, classes such as run and billiards rarely occur together in the same frame. 


MultiTHUMOS is a very challenging dataset for four key reasons.

\begin{enumerate}
    \item {\bf Long tail data distribution.} First, MultiTHUMOS has a long tail distribution in the amount of annotated data per action class. This requires action detection algorithms to effectively utilize both small and large amounts of annotated data. Concretely, MultiTHUMOS has between 27 seconds to 5 hours of annotated video per action class (with the rarest actions being volleyball bump, a pat, volleyball serve, high five and basketball block, and the most common actions being stand, walk, run, sit and talk to the camera). In contrast, THUMOS is more uniformly annotated: the dataset ranges from the rarest action baseball pitch with 3.7 minutes annotated to the most common action pole vault with 1 hour of annotated video. Figure~\ref{fig:hist-training-data} shows the full distribution. 
    \item {\bf Length of actions.} The second challenge is that MultiTHUMOS has much shorter actions compared to THUMOS. For each action class, we compute the average length of an action instance of that class. Instance of action classes in THUMOS are on average 4.8 second long compared to only 3.3 seconds long in MultiTHUMOS. Instances of action classes in THUMOS last between 1.5 seconds on average for clicket bowling to 14.7 seconds on average for billiards. In contrast, MultiTHUMOS has seven action classes whose instances last less than a second on average: two-handed catch, planting the pole in pole vaulting, basketball shot, one-handed catch, basketball block, high five and throw. Shorter actions are more difficult to detect since there is very little visual signal in the positive frames. There are instances of actions throw, body contract and squat that last only 2 \emph{frames} (or 66 milliseconds) in MultiTHUMOS! Accurately localizing such actions encourages  strong contextual modeling and multi-action reasoning.
   \item {\bf  Fine-grained actions.} The third challenge of MultiTHUMOS is the many fine-grained action categories with low visual inter-class variation, including hierarchical (e.g. throw vs. baseball pitch), hierarchical within a sport (e.g. pole vault vs. the act of planting the pole when pole vaulting), and fine-grained (e.g. basketball dunk, shot, dribble, guard, block, and pass). It also contains both sport-specific actions (such as different basketball or volleyball moves), as well as general actions that can occur in multiple sports (e.g. pump fist, or one-handed catch). This requires the development of general action detection approaches that are able to accurate model a diverse set of visual appearances.
\item {\bf High intra-class variation.} The final MultiTHUMOS challenge is the high intra-class variation as shown in Figure~\ref{fig:intra-class-variation}. The same action looks visually very different across multiple frames. For example,  a hug can be shown from many different viewpoints, ranging from extreme close-up shots to zoomed-out scene shots, and may be between two people or a larger group. This encourages the development of models that are insensitive to particular camera viewpoint and instead accurately focus on the semantic information within a video.
\end{enumerate}

With the MultiTHUMOS dataset providing new challenges for action detection, we now continue on to describing our proposed approach for addressing these challenges and making effective use of the dense multilabel annotation.




\section{Technical Approach}


Actions in videos exhibit rich patterns, both within a single frame due to action label relations and also across frames as they evolve in time. The desire to elegantly incorporate these cues with state-of-the-art appearance-based models has led to recent works \cite{donahue2014long,ng2015beyond,srivastava2015unsupervised} that study combinations of Convolutional Neural Networks (CNN) modeling frame-level spatial appearance and Recurrent  Neural Networks (RNN) modeling the temporal dynamics. However, the density of the action labels in our dataset expands the opportunities for more complex modeling at the temporal level. While in principle even a simple instantiation of an ordinary RNN has the capacity to capture arbitrary temporal patterns, it is not necessarily the best model to use in practice. Indeed, our proposed MultiLSTM model extends the recurrent models described in previous work, and our experiments demonstrate its effectiveness.

\subsection {LSTM}
\label{sec:lstm}


The specific type of Recurrent architecture that is commonly chosen in  previous work is the Long Short-Term Memory (LSTM), which owing to its appealing functional properties has brought success in a wide range of sequence-based tasks such as speech recognition, machine translation and very recently, video activity classification. Let $\mathbf{x}$ be an input sequence $(x_1,...,x_T)$ and $\mathbf{y}$ be an output sequence $(y_1,...,y_T)$. An LSTM then maps $\mathbf{x}$ to $\mathbf{y}$ through a series of intermediate representations:
\begin{align}
i_t &= \sigma(W_{xi}x_t + W_{h_i}h_{t-1} + b_i)\\
f_t &= \sigma(W_{xf}x_t + W_{hf}h_{t-1} + b_f)\\
o_t &= \sigma(W_{xo}x_t + W_{ho}h_{t-1} + b_o)\\
g_t &= \tanh(W_{xc}x_t + W_{hc}h_{t-1} + b_c)\\
c_t &= {f_t}{c_{t-1}} + {i_t}{g_t}\\
h_t &= {o_t}\tanh(c_t)\\
y_t &= W_{hy}h_t + b_y
\end{align}
Here $c$ is the ``internal memory'' of the LSTM, and the gates $i$, $f$, $o$ control the degree to which the memory accumulates new input $g$, attenuates its memory, or influences the hidden layer output $h$, respectively. Intuitively, the LSTM has the capacity to read and write to its internal memory, and hence maintain and process information over time. Compared to standard RNNs, the LSTM networks mitigate the ``vanishing gradients'' problem because except for the forget gate, the cell memory is influenced only by additive interactions that can communicate the gradient signal over longer time durations. The architecture is parametrized by the learnable weight matrices $W$ and biases $b$\rebuttal{, and we refer the reader to \cite{HochreiterS97,donahue2014long} for further details.}

However, an inherent flaw of the plain LSTM architecture is that it is forced to make a definite and final prediction at some time step based on what frame it happens to see at that time step, and its previous context vector.

\subsection {MultiLSTM}
\label{sec:multilstm}

Our core insight is that providing the model with more freedom in both reading its input and writing its output reduces the burden placed on the hidden layer representation. Concretely, the MultiLSTM expands the temporal receptive field of both input and output connections of an LSTM. These allow the model to directly refine its predictions in retrospect after seeing more frames, and additionally provide direct pathways for referencing previously-seen frames without forcing the model to maintain and communicate this information through its recurrent connections.

\subsubsection{Multilabel Loss}

In our specific application setting, the input vectors $x_t$ correspond to the 4096-dimensional fc-7 features of the VGG 16-layer Convolutional Network which was first pretrained on ImageNet and then fine-tuned on our dataset on an individual frame level. We interpret the vectors $y_t$ as the unnormalized log probability of each action class. Since each frame of a video can be labeled with multiple classes, instead of using the conventional softmax loss we sum independent logistic regression losses per class:
\begin{equation*}
L(\mathbf{y} | \mathbf{x}) = \sum_{t,c} z_{tc}\log(\sigma(y_{tc})) + (1-z_{tc})\log(1-\sigma(y_{tc}))
\end{equation*}
where $y_{tc}$ is the score for class $c$ at time $t$, and $z_{tc}$ is the binary ground truth label for class $c$ at time $t$. 

\begin{figure}
\begin{tabular}{@{}l@{}l@{}}
\multicolumn{2}{c}{\includegraphics[width=0.95\linewidth]{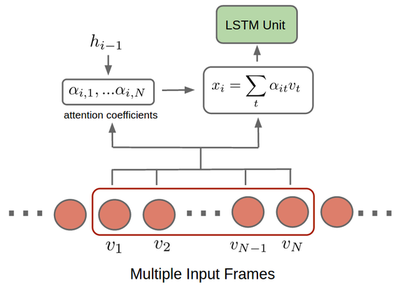}} \\
\multicolumn{2}{p{\linewidth}}{(a) Connections to multiple inputs.}  \\
\multicolumn{2}{l}{\includegraphics[width=0.95\linewidth]{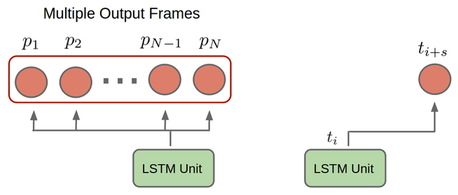}}\\
\multicolumn{1}{r}{(b) Multiple outputs.} &
\multicolumn{1}{r}{(c) Variant: output offset.}
\end{tabular}
\caption{Components of our MultiLSTM model.}
\label{fig:lstm-atten} 
\end{figure}

\subsubsection{Multiple Inputs with Temporal Attention}

In a standard LSTM network, all contextual information is summarized in the hidden state vector.  Therefore, the network relies on the memory vector to contain all relevant information about past inputs, without any ability to explicitly revisit past inputs. This is particularly challenging in the context of more complex tasks such as dense, multilabel action detection.

To provide the LSTM with a more direct way of accessing recent inputs, we expand the temporal dimension of the input to be a fixed-length window of frames previous to the current time step (Figure~\ref{fig:lstm-atten}(a)). This allows the LSTM to spend its modeling capacity on more complex and longer-term interactions instead of maintaining summary of the recent frames in case it may be useful for the next few frames. Furthermore, we incorporate a soft-attention weighting mechanism that has recently been proposed in the context of machine translation \cite{bahdanau2014neural}.

Concretely, given a video $\mathbf{V} = \{v_1, \dots v_T\}$, the input $x_i$ to the LSTM at time step $i$ is now no longer the representation of a single frame $v_t$, but a weighted combination $x_i = \sum_{t} \alpha_{it}{v_t}$ where $t$ ranges over a fixed-size window of frames previous to $i$, and $\alpha_{it}$ is the contribution of frame $v_t$ to input $x_i$ as computed by the soft attention model. To compute the attention coefficients $\alpha_{it}$, we use a model similar to Bahdanau et al.~\cite{bahdanau2014neural}. The precise formulation that worked best in our experiments is:
\begin{equation}
\alpha_{it} \propto  \exp(w_{ae}^T \left[ \tanh(W_{ha}h_{i-1}) \odot \tanh(W_{va}v_t) \right]) 
\end{equation}
Here $\odot$ is element-wise multiplication, \rebuttal{$\{w_{ae}, W_{ha}, W_{va}\}$ are learned weights}, and $\alpha_t$ is normalized using the softmax function with the interpretation that $\alpha_t$ expresses the relative amount of attention assigned to each frame in the input window. Intuitively, the first term $\tanh(W_{ha}h_{i-1})$ allows the network to look for certain features in the input, while the second term $\tanh(W_{va}v_t)$ allows each input to broadcast the presence/absence of these features. Therefore, the multiplicative interaction followed by the weighted sum with $w_{ae}$ has the effect of quantifying the agreement between what is present in the input and what the network is looking for. Note that the standard LSTM formulation is a special case of this model where all attention is focused on the last input window frame.

\subsubsection{Multiple Outputs}

Analogous to providing explicit access to a window of frames at the input, we allow the LSTM to contribute to predictions in a window of frames at the output (Figure \ref{fig:lstm-atten}(b)). Intuitively, this mechanism lets the network refine its predictions in retrospect, after having seen more frames of the input. This feature is related to improvements that can be achieved by use of bi-directional recurrent networks. However, unlike bi-directional models our formulation can be used in an online setting where it delivers immediate predictions that become refined with a short time lag.\footnote{A similar behavior can be obtained with a bi-directional model by truncating the hidden state information from future time frames to zero, but this artificially distorts the test-time behavior of the model's outputs, while our model always operates in the regime it was trained with.} Given the multiple outputs, we consolidate the predicted labels for all classes $c$ at time $t$ with a weighted average $y_{t} = \sum_{i} \beta_{it}p_{it}$ where $p_{it}$ are the predictions at the $i$th time step for the $t$th frame, and $\beta_{it}$ weights the contribution. $\beta_{it}$ can be learned although in our experiments we use $\frac{1}{N}$ for simplicity to average the predictions. The standard LSTM is a special case, where $\beta$ is an indicator function at the current time step. 
In our experiments we use the same temporal windows at the input and output. Similar to the inputs, we experimented with soft attention over the output predictions but did not observe noticeable improvements. \revision{This may be due to increased fragility when the attention is close to the output without intermediate network layers to add robustness, and we leave further study of this to future work.}


\subsubsection{Single Offset Output}

We experimented with offset predictions to quantify how informative frames at time $t$ are towards predicting labels at some given offset in time. In these experiments, the network is trained with shifted labels $y_{t+s}$, where $s$ is a given offset (Figure \ref{fig:lstm-atten}(c)). In our dense label setting, this type of model additionally enables applications such as action prediction in unconstrained internet video (c.f.~\cite{KitaniZBH12}).  For example, if the input is a frame depicting a person cocking his arm to throw, the model could predict future actions such as Catch or Hit.


\section {Experiments}

We begin by describing our experimental setup in Section~\ref{sec:expsetup}. We then empirically demonstrate the effectiveness of our model on the challenging tasks of action detection (Section~\ref{sec:exp_action_det}) and action prediction (Section~\ref{sec:expprediction}).

\subsection{Setup}
\label{sec:expsetup}

\subsubsection{Dataset}
We evaluate our MultiLSTM model for dense, multilabel action detection on the MultiTHUMOS dataset. \rebuttal{We use the same train and test splits as THUMOS (see Sec. \ref{sec:dataset} for details) but ignore the background training videos.  Clipped training videos (the ``Training Data" set in THUMOS) act as weak supervision since they are only labeled with the THUMOS-subset of MultiTHUMOS classes.}

\subsubsection{Implementation Details} Our single-frame baseline uses the 16-layer VGG CNN model \cite{Simonyan14c}, which achieves near state of the art performance on ILSVRC~\cite{ILSVRC}. The model was pre-trained on ImageNet and \revision{all layers fine-tuned on MultiTHUMOS using a binary cross-entropy loss per-class.} The input to our LSTM models is the final 4096-dimensional, frame-level fc7 representation. 

We use 512 hidden units in the LSTM, and \revision{50 units in the attention component of MultiLSTM that is used to compute attention coefficients over a window of 15 frames.}  We train the model with an exact forward pass, passing LSTM hidden and cell activations from one mini-batch to the next. However we use approximate backpropagation through time where we only backpropagate errors for the duration of a single mini-batch. Our mini-batches consist of 32 input frames (approx. 3.2 seconds), and we use RMSProp~\cite{rmsprop} to modulate the per-parameter learning rate during optimization.

\subsubsection{Performance Measure}
We evaluate our models using Average Precision (AP) measured on our frame-level labels. The focus of our work is dense labeling, hence this is the measure we analyze to evaluate the performance of our model.  We report AP values for individual action classes as well as mean Average Precision (mAP), the average of these values across the action categories.

To verify that our baseline models are strong, we can obtain discrete detection instances using standard heuristic post-processing.  Concretely, for each class we threshold the frame-level confidences at $\lambda$ ($\lambda = 0.1$ obtained by cross-validation) to get binary predictions and then accumulate consecutive positive frames into detections. For each class $C$, let $\mu(C)$ and $\sigma(C) $ be the mean and standard deviation respectively of frame lengths on the training set. The score of a detection for class $C$ of length $L$ with frame probabilities $p_1 \dots p_L$ is then computed as
\begin{equation}
   score(C, p_1 \dots p_L) =  (\sum_i^L p_i)\times \exp(\frac{-\alpha (L-\mu(C))^2}{\sigma(C)^2})
\end{equation}
where the hyperparameter $\alpha = 0.01$ is obtained by cross-validation.  Using this post-processing, our single-frame CNN model achieves 32.4 detection mAP with overlap threshold $0.1$ on the THUMOS subset of MultiTHUMOS. Since state of the art performance on THUMOS reports 36.6 detection mAP including audio features, this confirms that our single-frame CNN is a reasonable baseline. Hereafter, we compare our models without this post-processing to achieve a comparison of the models' dense labeling representational ability.


\begin{table}

\centering
\begin{tabular}{|l|c|c|}
\hline
Model & THUMOS mAP  & MultiTHUMOS mAP \\
\hline
\revision{IDT~\cite{Wang2013}} & \revision{13.6} & \revision{13.3}\\
Single-frame CNN~\cite{Simonyan14c} & 34.7 & 25.4 \\
Two-stream CNN~\cite{simonyan2014two} & 36.2 & 27.6 \\
LSTM & 39.3 & 28.1 \\
LSTM + i & 39.5 & 28.7 \\
LSTM + i + a & 39.7 & 29.1 \\
MultiLSTM & {\bf 41.3} & {\bf 29.7} \\
\hline
\end{tabular}
\caption{Per-frame mean Average Precision (mAP) of the MultiLSTM model compared to baselines. Two-stream CNN is computed with single-frame flow. LSTM is implemented in the spirit of~\cite{donahue2014long} (details in Section~\ref{sec:multilstm}). We show the relative contributions of adding first the input connections with averaging (LSTM + i), then the attention (LSTM + i + a) as in Figure~\ref{fig:lstm-atten}(a), and finally the output connections to create our proposed MultiLSTM model (LSTM + i + a + o) as in Figure~\ref{fig:lstm-atten}(b).}
\label{tab:frame_ap}
\end{table}


\begin{figure}
\centering
\begin{tabular}{@{}l@{}}
\includegraphics[width=0.9\linewidth]{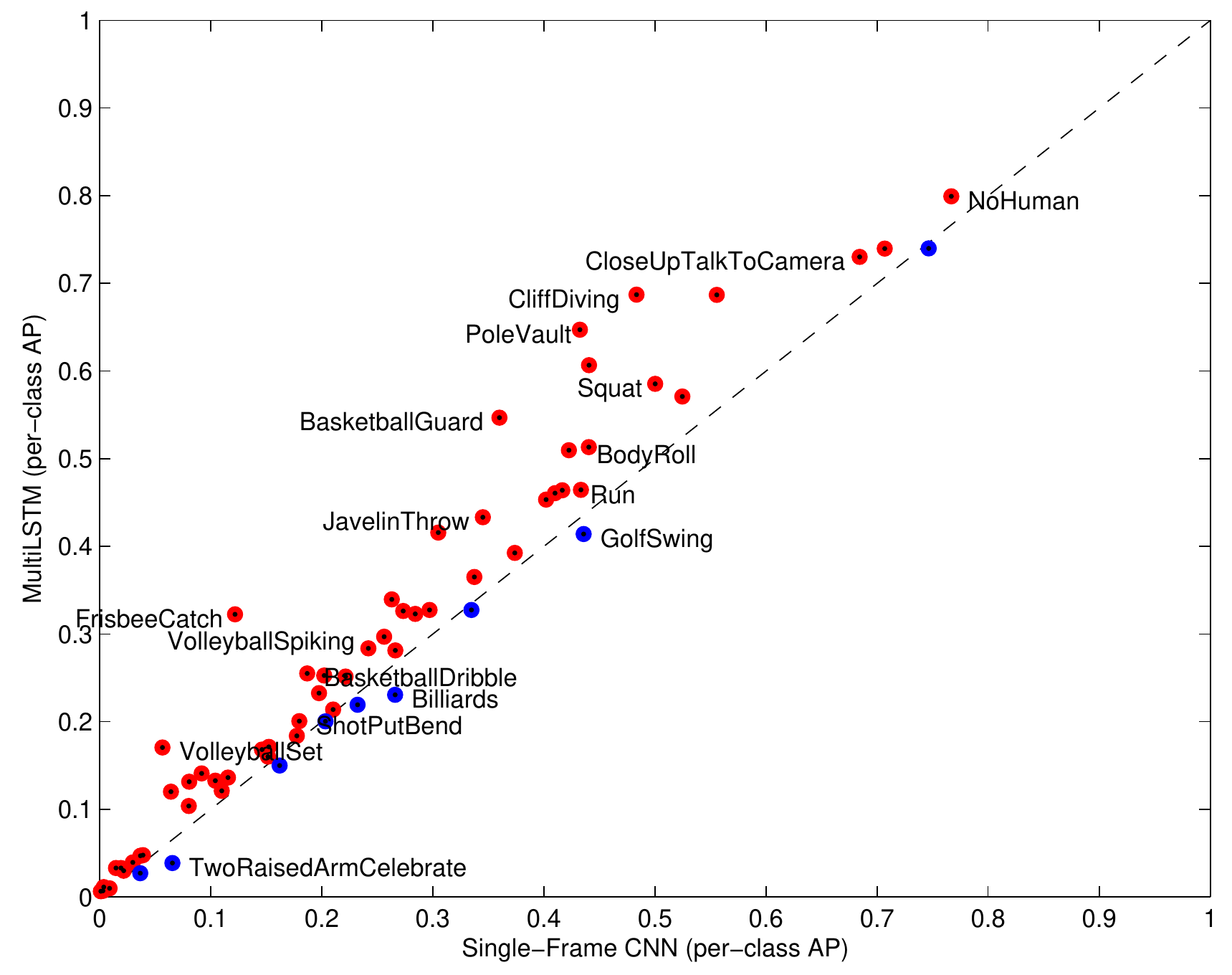}\\
(a) Single-frame CNN vs. MultiLSTM \\
\includegraphics[width=0.9\linewidth]{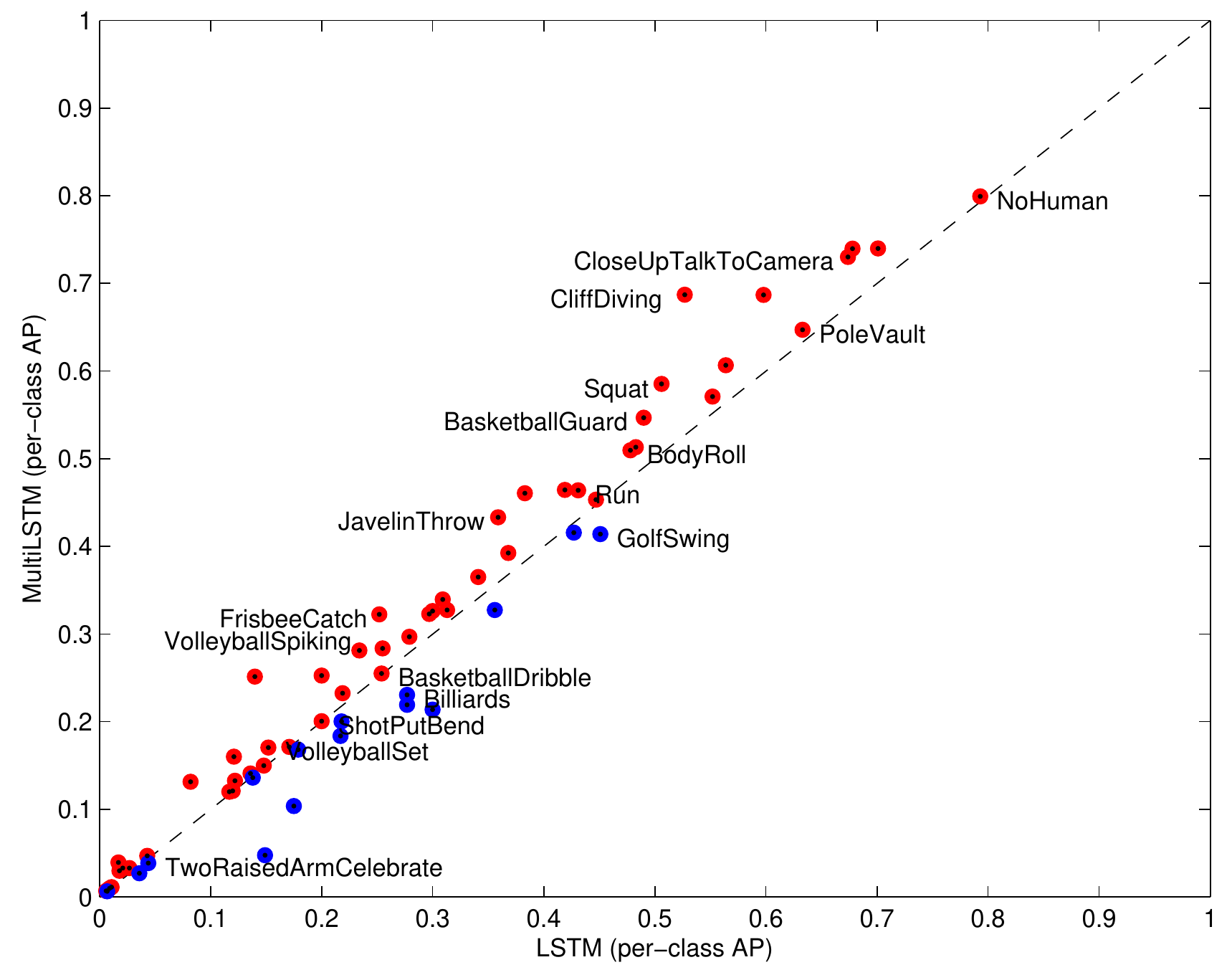}\\
(b) LSTM vs. MultiLSTM \\
\end{tabular}
\caption{\revision{Per-class Average Precision of the MultiLSTM  model compared to (a) a single-frame CNN model \cite{Simonyan14c}; and (b) an LSTM on MultiTHUMOS. MultiLSTM outperforms the single-frame CNN on 56 out of 65 action classes, and the LSTM on 50 out of 65 action classes.}}
\label{fig:frame-ap}
\end{figure}

\begin{figure}
\centering
\includegraphics[width=0.6\linewidth]{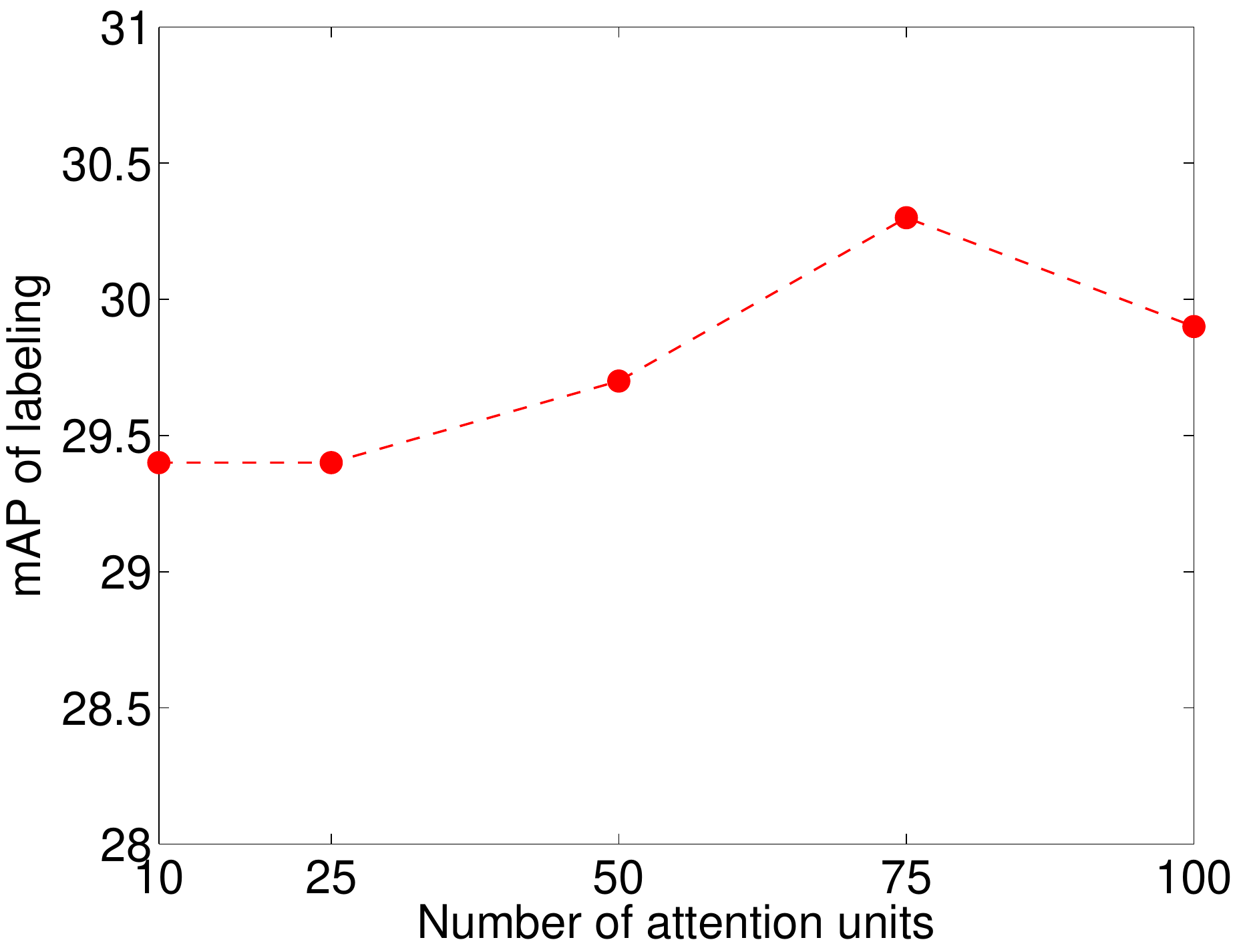} \\
\caption{\revision{Number of attention units vs. per-frame mAP of the MultiTHUMOS model. Performance increases as the number of units is increased, but decreases past 75 units. We use 50 units in our experiments.}}
\label{fig:attention-units}
\end{figure}


\begin{figure*}
\centering
\includegraphics[width=0.9\linewidth]{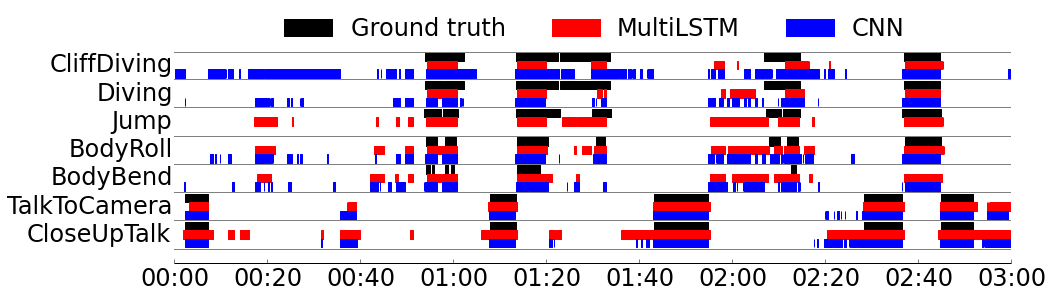}
\caption{Example timeline of multilabel action detections from our MultiLSTM model compared to a CNN. (best in color)}
\label{fig:timeline}
\end{figure*}

\subsection{Action Detection}
\label{sec:exp_action_det}

We first evaluate our models on the challenging task of dense per-frame action labeling on MultiTHUMOS. The MultiLSTM model achieves consistent improvements in mean average precision (mAP) compared to baselines. \revision{A model trained on Improved Dense Trajectories features~\cite{Wang2013} (using a linear SVM trained on top of a temporally pooled and quantized dictionary of pre-computed IDT features, provided by THUMOS'14) performs relatively poorly with 13.3 mAP.  This highlights the difficulty of the dataset and the challenge of working with generic hand-crafted features that are not learned for these specific fine-grained actions.} \revision{Additional variants of IDT could be used to improve performance.  For example, Fisher Vector encoding of raw IDT features is commonly used to boost performance.  However, these methods can be computationally expensive and are limited due to their reliance on underlying hand-crafted features and lack of opportunity for joint training.  Hence, we use neural network-based models for the rest of our experiments.} 

A single-frame CNN fine-tuned on MultiTHUMOS attains $25.4\%$ mAP. We trained a base LSTM network in the spirit of \cite{donahue2014long} but modified for multilabel action labeling.  Specifically, the LSTM is trained using a multilabel loss function and tied hidden context across 32 frame segments, as described in Section~\ref{sec:multilstm}.  This base LSTM boosts mAP to $28.1\%$. Our full MultiLSTM model handily outperforms both baselines with $29.7\%$ mAP. Table~\ref{tab:frame_ap} additionally demonstrates that each component of our model (input connections, input attention and output connections) is important for accurate action labeling.

\revision{Figure~\ref{fig:frame-ap} compares per-class results of the CNN vs. MultiLSTM, and the base LSTM vs. MultiLSTM. MultiTHUMOS outperforms the CNN on 56 our of 65 action classes, and the LSTM on 50 out of 65 action classes. A sampling of action classes is labeled. It is interesting to note from the two plots that compared with the CNN, the LSTM closes the gap with MultiLSTM on classes such as Frisbee Catch, Pole Vault, and Basetkball Guard, which are strongly associated with temporal context (e.g. a throw proceeds a frisbee catch, and a person usually stands at the track for some time before beginning a pole vault). This shows the benefit of stronger temporal modeling, which MultiLSTM continues to improve on the majority of classes.}

\revision{Figures~\ref{fig:attention-units} analyzes per-frame mAP as the number of attention units (at both input and output) in the MultiLSTM model is varied. We observe that increasing the number of attention units improves performance up to a point (75 units), as would be expected, and then decreases past that as the number of parameters becomes too large. In practice, we use 50 units in our experiments.}

Figure~\ref{fig:timeline} visualizes some results of MultiLSTM compared to a baseline CNN.
For ease of visualization, we binarize outputs by thresholding
rather than showing the per-frame probabilistic action labels our
model produces. The CNN often produces short disjoint detections whereas MultiLSTM effectively makes use of temporal and co-occurrence context to produce more consistent detections.

\begin{figure*}[ht]
\begin{tabular}{|c|c|}
\hline
\color{ForestGreen} { Pass, then Shot} & 
 \color{ForestGreen} { Pass, then Shot} \\
 \includegraphics[width=0.48\linewidth]{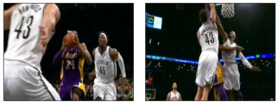} &
   \includegraphics[width=0.48\linewidth]{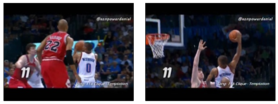} \\
\hline
 \color{ForestGreen} { Jump, then Fall} &
  \color{ForestGreen} { Jump, then Fall} \\
   \includegraphics[width=0.48\linewidth]{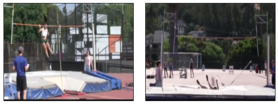} &
      \includegraphics[width=0.48\linewidth]{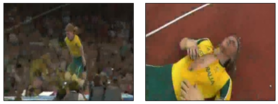}\\
\hline
\color{ForestGreen} { Throw, then OneHandedCatch} &
 \color{ForestGreen} { Throw, then TwoHandedCatch} \\
\includegraphics[width=0.48\linewidth]{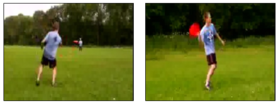} &
 \includegraphics[width=0.48\linewidth]{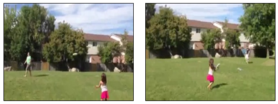} \\
\hline
  \color{BrickRed}   { Clean, then Jerk}  &
 \color{BrickRed}  { Pitch, then OneHandedCatch} \\
     \includegraphics[width=0.23\linewidth]{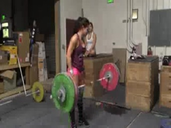} 
  \includegraphics[width=0.23\linewidth]{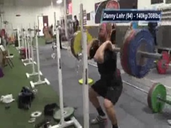} &
  \includegraphics[width=0.23\linewidth]{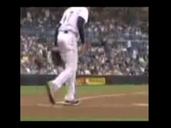} 
  \includegraphics[width=0.23\linewidth]{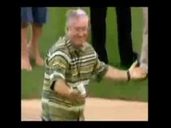} \\
\hline
\end{tabular}
\caption{Examples of retrieved sequential actions (correct in green, mistakes in red).  Results are shown in pairs: first action frame on the left, second action frame  on the right.}
\label{fig:ret-seq}
\end{figure*}
\begin{figure*}
\begin{tabular}{|cc|cc|}
\hline
\color{ForestGreen} { Shot\&Guard }&
\color{ForestGreen} { Shot\&No Guard} &
\color{ForestGreen} {  Sit\&Talk }&
\color{ForestGreen} { Stand\&Talk }\\
  	\includegraphics[width=0.23\linewidth]{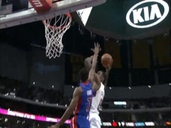} &
  	\includegraphics[width=0.23\linewidth]{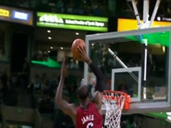} &
  	\includegraphics[width=0.23\linewidth]{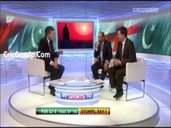} &
\includegraphics[width=0.23\linewidth]{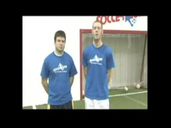} \\
\hline
   	\color{ForestGreen} { Dive\&Bodyroll} &
\color{ForestGreen}{  Dive\&No Bodyroll} &
\color{BrickRed} { Hug\&Pat} &
\color{BrickRed} { PlantPole\&Run} \\
\includegraphics[width=0.23\linewidth]{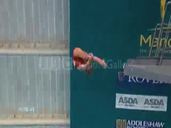} &
    	\includegraphics[width=0.23\linewidth]{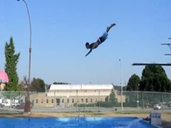} &
    	\includegraphics[width=0.23\linewidth]{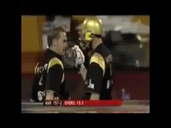} &
    	\includegraphics[width=0.23\linewidth]{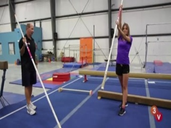} \\
    	\hline
\end{tabular}
\caption{Examples of retrieved frames with co-occurring actions (correct in green, mistakes in red). The model is able to distinguish between subtly different scenarios. %
}
\label{fig:ret-coc}
\end{figure*}

The multilabel nature of our model and dataset allows us to go beyond simple action labeling and tackle higher-level tasks such as retrieval of video segments containing sequences of actions (Figure~\ref{fig:ret-seq}) and co-occurring actions (Figure~\ref{fig:ret-coc}). By learning accurate co-occurrence and temporal relationships, the model is able to retrieve video fragments with detailed action descriptions such as Pass and then Shot or frames with simultaneous actions such as Sit and Talk.


\begin{figure}
\centering
\begin{tabular}{@{}c@{}c@{}}
\includegraphics[width=0.5\linewidth]{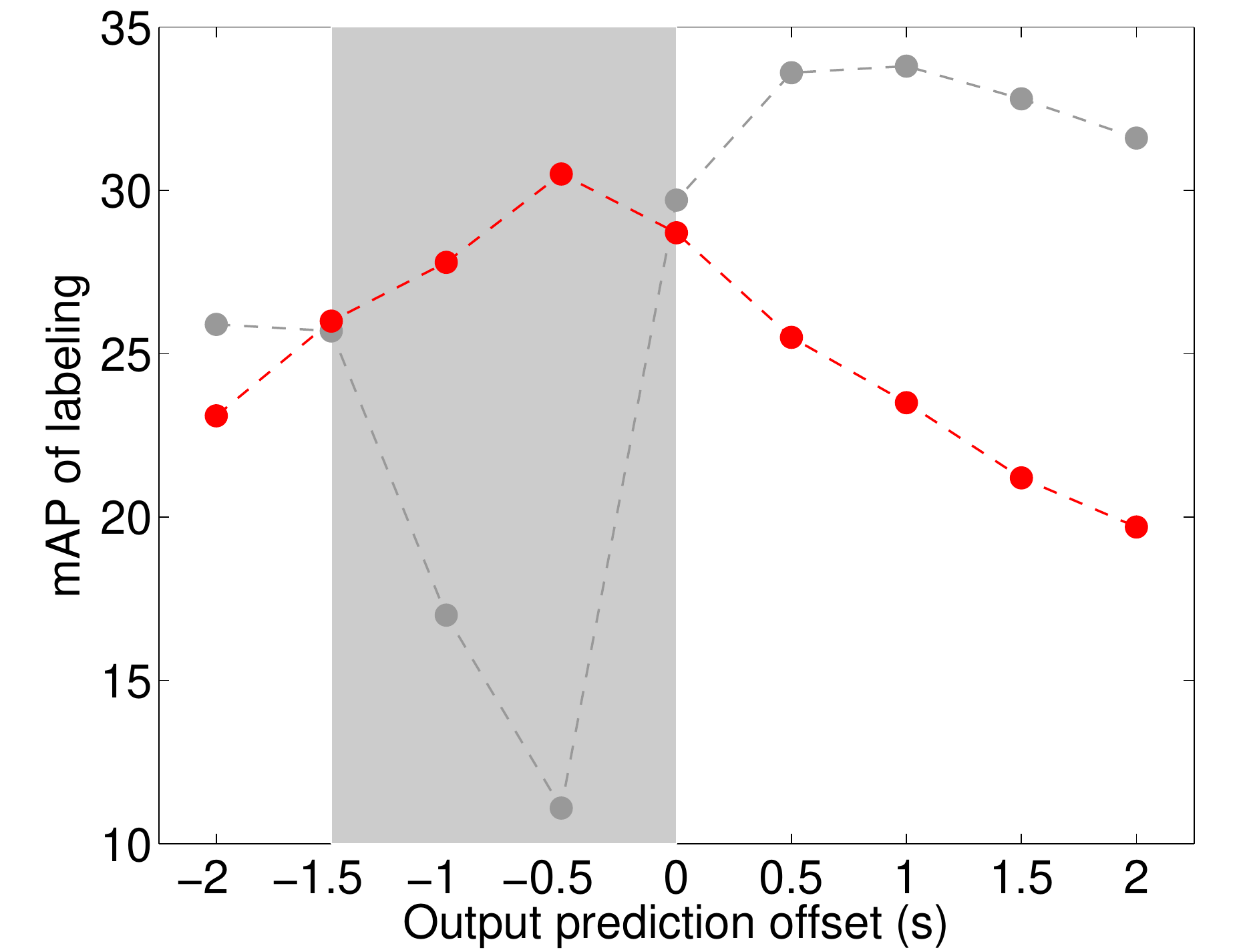} &
\includegraphics[width=0.5\linewidth]{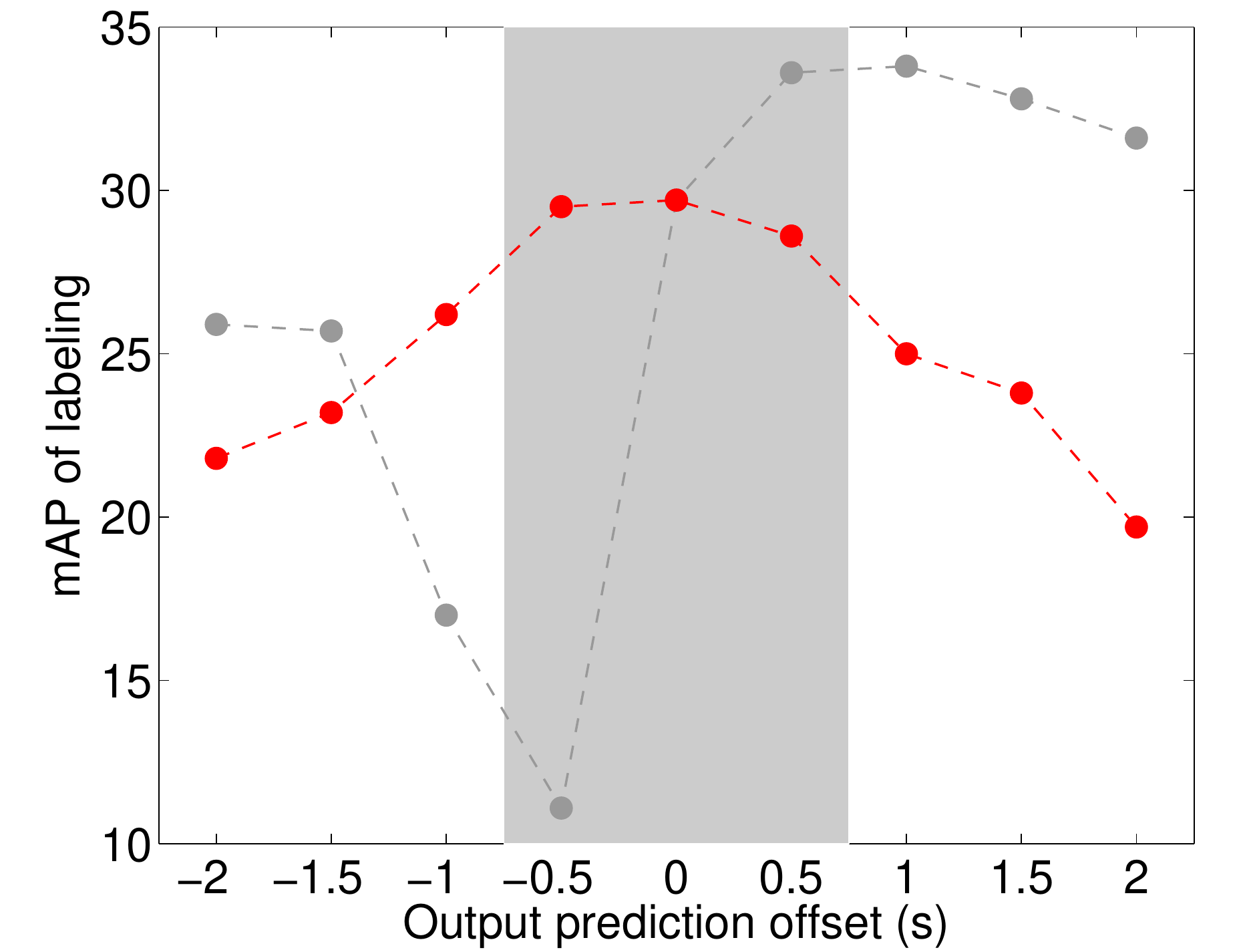} \\
\end{tabular}
\caption{\revision{Action detection mAP when the MultiLSTM model predicts the action for a past (offset $<$ 0) or future (offset $>$ 0) frame rather than for the current frame (offset $=$ 0). The input window of the MultiLSTM model is shown in gray. Thus, the left plot is of a model trained with input from the past, and the right plot is of a model trained with the input window centered around the current frame. mAP of the MultiLSTM model is shown in red, and mAP of a model using ground-truth label distribution is shown in gray.}}
\label{fig:input_shifts}
\end{figure}

\begin{figure*}
\centering
\begin{tabular}{|c|c|}
\hline
\color{ForestGreen}{Jump $\rightarrow$ Fall}&
\color{ForestGreen}{Jump $\rightarrow$ Fall}
\\
\includegraphics[width=0.48\linewidth]{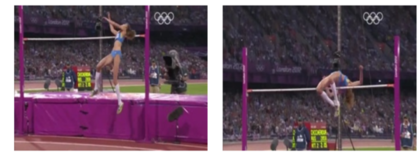} &
\includegraphics[width=0.48\linewidth]{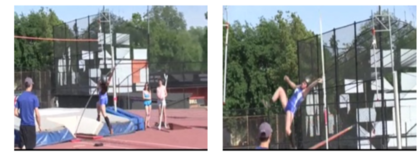}
\\
\hline
\color{ForestGreen}{Dribble $\rightarrow$ Shot} &
  \color{ForestGreen}{Dribble $\rightarrow$ Shot}
\\
 \includegraphics[width=0.48\linewidth]{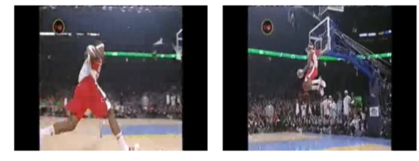} &
 \includegraphics[width=0.48\linewidth]{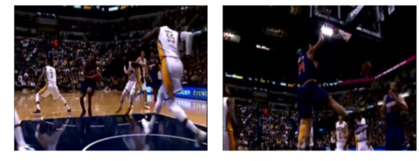}
 \\
 \hline
\color{ForestGreen} {DiscusWindUp $\rightarrow$ Release}& 
\color{ForestGreen} {DiscusWindUp $\rightarrow$ Release}
\\
\includegraphics[width=0.48\linewidth]{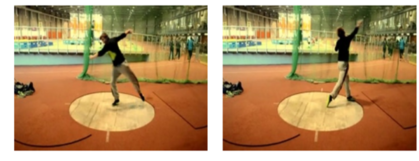} &
\includegraphics[width=0.48\linewidth]{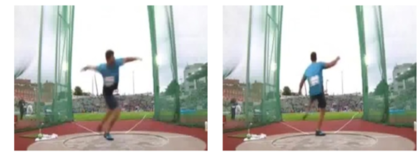}
\\
\hline
\color{ForestGreen} {VolleyballServe $\rightarrow$ VolleyballSpiking}& 
\color{ForestGreen} {VolleyballServe $\rightarrow$ VolleyballSpiking}
\\
\includegraphics[width=0.48\linewidth]{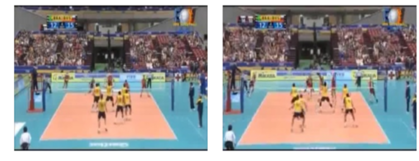} &
\includegraphics[width=0.48\linewidth]{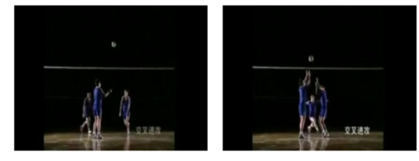}
\\
\hline
\color{BrickRed} Dribble $\rightarrow$ Shot &
  \color{BrickRed} Jump $\rightarrow$ Fall\\   
         \includegraphics[width=0.48\linewidth]{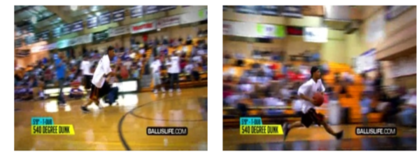} &
         \includegraphics[width=0.48\linewidth]{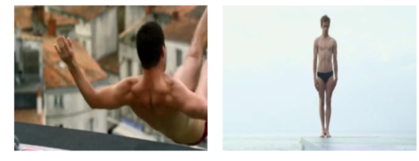} \\    \hline
         
 \end{tabular}
\caption{Examples of predicted actions.  For each pair of actions, the first one (left) is the label of the current frame and the second one (right) is the predicted label 1 second into the future. Correct predictions are shown in green, and failure cases are shown in red.}
\label{fig:temp-atten-examples}
\end{figure*}

\subsection{Action Prediction}
\label{sec:expprediction}

Dense multilabel action labeling in unconstrained internet videos is a challenging problem to tackle in and of itself. In this section we go one step further and aim to make predictions about what is likely to happen next or what happened previously in the video. By utilizing the MultiLSTM model with offset (Figure~\ref{fig:lstm-atten}(c)) we are able to use the learned temporal relationships between actions to make inferences about actions likely occurring in past or future frames.

We evaluate the performance of this model as a function of temporal offset magnitude and report results in Figure \ref{fig:input_shifts}. \revision{MultiLSTM prediction mAP is shown in red.} The plot on the left quantifies the prediction ability of the model within a 4 second (+/- 2 second) window, provided an input window of context spanning the previous 1.5 seconds.  The model is able to ``see the future'' -- while predicting actions 0.5 seconds in the past is easiest (mAP $\approx 30\%$), reasonable prediction performance (mAP $\approx 20-25\%$) is possible 1-2 seconds into the future. The plot on the right shows the prediction ability of the model using an input context centered around the current frame, instead of spanning only the past.  The model is able to provide stronger predictions at past times compared to future times, giving quantitative insight into the contribution of the hidden state vector to providing past context.

\revision{It is also interesting to compare MultiLSTM prediction to a model using the ground-truth label distribution (shown in gray). Specifically, this model makes action predictions using the most frequent label for a given temporal offset from the training set, per-class, and weighted by the MultiLSTM prediction probabilities of actions in the current frame. The label distribution-based model has relatively high performance in the future direction as opposed to the past, and at farther offsets from the current frame. This indicates that stronger priors can be learned in these temporal regions (e.g. frisbee throw should be followed by frisbee catch, and 2 seconds after a dive is typically background (no action)), and MultiLSTM does learn them to some extent. On the other hand, the label distribution-based model has poor performance immediately before the current frame, indicating that there is greater variability in this temporal region, e.g. clapping may be preceded by many different types of sport scoring actions, though a longer offset in the past may be more likely background. In this temporal region, MultiLSTM shows significantly stronger performance than using priors, indicating the benefit of its temporal modeling in this context.} 

Figure~\ref{fig:temp-atten-examples} shows qualitative examples of predictions at frames 1 second in the future from the current time.  The model is able to correctly infer that a Fall is likely to happen after a Jump, and a BasketballShot soon after a Dribble. 


\section {Conclusion}

In conclusion, this paper presents progress in two aspects of human action understanding.  First, we emphasize a broader definition of the task, reasoning about dense, multiple labels per frame of video.  We have introduced a new dataset MultiTHUMOS, containing a substantial set of labeled data that we will release to spur research in this direction of action recognition.  Second, we develop a novel LSTM-based model incorporating soft attention input-output temporal context for dense action labeling.  We show that utilizing this model on our dataset leads to improved accuracy of action labeling and permits detailed understanding of human action.

\section*{Acknowledgments}
We would like to thank Andrej Karpathy and Amir Zamir for helpful comments and discussion.

{\small
\bibliographystyle{ieee}
\bibliography{ijcv2016}
}

\end{document}